# SINGULARITY ANALYSIS OF LIMITED-DOF PARALLEL MANIPULATORS USING GRASSMANN-CAYLEY ALGEBRA

Daniel Kanaan, Philippe Wenger, Damien Chablat
*Institut de Recherche en Communication et Cybernétique de Nantes*
*1, rue de la Noë, 44321 Nantes, FRANCE*
[Daniel.kanaan, Philippe.wenger, Damien.chablat]@irccyn.ec-nantes.fr

**Abstract**    This paper characterizes geometrically the singularities of limited DOF parallel manipulators. The geometric conditions associated with the dependency of six Plücker vector of lines (finite and infinite) constituting the rows of the inverse Jacobian matrix are formulated using Grassmann-Cayley algebra. Manipulators under consideration do not need to have a passive spherical joint somewhere in each leg. This study is illustrated with three example robots.

## 1. Introduction

Parallel singularities are critical configurations in which a parallel manipulator loses its stiffness and gains one and more degrees of freedom (DOF). These singular configurations can be found using analytical, numerical or geometrical methods (Merlet, 2005). The inverse Jacobian matrix of a 6-DOF parallel manipulator has Plücker coordinate vectors of finite lines as its rows. These lines are wrenches of actuation describing the instantaneous forces of actuation applied by the actuators on the moving platform. For parallel manipulators of type Gough-Stewart, parallel singularities occur when lines within legs of the manipulator are linearly dependent, which Merlet, 1989, analyzed using Grassmann line geometry. Hao and McCarthy, 1998, have used screw theory to define conditions for line-based singularities. They focused on 6-DOF parallel manipulators, each leg having at least one actuator and their last three joints equivalent to a passive spherical joint. These conditions ensure that each supporting leg of the system can apply only a pure force to the platform so that it is possible to characterize all singular configurations in terms of the geometry of linearly dependent sets of lines. Ben-Horin and Shoham, 2005 and 2006 analyzed parallel singularities of two classes of 6-DOF parallel manipulators using

Grassmann-Cayley algebra (GCA). They used the superbracket and the Grassmann-Cayley operators to obtain geometric conditions for singularity, namely, when four planes defined by the direction of the joint axes and the location of the spherical joint are concurrent in a point.

Contrary to 6-DOF manipulators, each leg of a limited-DOF parallel manipulator has connectivity less than 6 and, in turn, constraints partly the motion of the moving platform. When a leg loses its ability to constrain the moving platform, a so-called constraint singularity occurs. Joshi and Tsai, 2002 developed a methodology for the inverse Jacobian of limited-DOF parallel manipulators with the theory of reciprocal screws. A 6x6-inverse Jacobian matrix was defined, the rows of which are wrenches that provide information about both architecture[1] and constraint singularities. These wrenches, also known as governing lines, are actuation and constraint wrenches applied to the moving platform.

In this paper, we focus on limited-dof parallel manipulators i.e. that have less than 6-DOF and we are interested in identifying their parallel singularities using GCA. The application of GCA is enlarged to include limited-DOF parallel manipulators, in which the inverse Jacobian is a 6x6 matrix having not only finite lines (zero pitch screws) as its rows but also infinite lines (infinite pitch screws). In this work, manipulators under consideration do not need to have a passive spherical joint somewhere in each of their legs. The results are illustrated with three example robots.

## 2. Grassmann-Cayley Algebra

Originally developed by H. Grassmann as a calculus for linear varieties, GCA has two operators, namely the join, denoted by $\vee$ and the meet, denoted by $\wedge$. These two operators are associated with union and intersection between vector subspaces of extensors. These extensors represent geometric entities such as points, lines, planes, etc. in the projective space. GCA makes it possible to work at the symbolic level, where points and lines are represented in a coordinate-free form by their homogeneous and Plücker coordinates, respectively.

GCA is used in this study to transform the singular geometric conditions defined as the dependency between six lines in the 3-dimensional projective space $\mathbb{P}^3$, into coordinate free algebraic expressions involving 12 points selected on the axes of these lines. For further details on GCA concept and its application to robotics see for example (White, 1994 and 2005, Staffetti and Thomas, 2000, Ben-Horin and Shoham, 2006) and references therein.

---

[1] Following Joshi and Tsai, 2002 terminology, an architecture singularity occurs when the inverse Jacobian matrix is rank-deficient and the Jacobian of constraints is full-rank.

### 2.1 Projective Space

The three-dimensional projective space $\mathbb{P}^3$ can be considered as the union of $\square^3$ with a set of ideal points that are the intersections of parallel lines and planes. The set of all such points forms a plane known as the plane at infinity, $\Pi_\infty$. This plane may be thought of as the set of all directions, since all lines with the same particular direction intersect $\Pi_\infty$ in the same unique point and all parallel planes intersect $\Pi_\infty$ in the same unique line. Hence every line at infinity meets every other line at infinity, and they are therefore all in one plane.

### 2.2 Bracket Ring

Consider a finite set of points $\{e_1, e_2,..,e_d\}$ defined in the *d*-dimensional vector space, *V*, over the field $\square$. Each point $e_i$ is represented by a *d*-tuple by using homogeneous coordinates, where $e_i=e_{1,i},e_{2,i},...,e_{d,i}$ (*1≤i≤d*). The bracket of these points is defined as the determinant of the matrix **M** whose columns are the homogeneous coordinates of these points $e_i$ (*1≤i≤d*):

$$[e_1,e_2,...,e_d] = \det \mathbf{M} = \begin{vmatrix} e_{1,1} & e_{1,2} & \cdots & e_{1,d} \\ \vdots & \vdots & \cdots & \vdots \\ e_{d,1} & e_{d,2} & \cdots & e_{d,d} \end{vmatrix} \quad (1)$$

The brackets form a subalgebra of the GCA, called the bracket ring or bracket algebra. The brackets satisfy the following relations:

1) $[e_1,e_2,...,e_d] = 0$ if any $e_i = e_j$ with $i \neq j$, or $e_1,e_2,...,e_d$ are dependent.
2) $[e_1,e_2,...,e_d]=sign(\sigma)[e_{\sigma 1},e_{\sigma 2},...,e_{\sigma d}]$ for any permutation σ of $\{1,2,...,d\}$
3) $[e_1,e_2,...,e_d][f_1,f_2,...,f_d]=\sum_{j=1}^{d}[f_j,e_2,...,e_d][f_1,f_2,...,f_{j-1},e_1,f_{j+1},...,f_d]$

All relations among the brackets are consequences of relations of the above three types. The relations of the third type are called Grassmann-Plücker relations or syzygies, and they correspond to generalized Laplace expansions by minors (White, 1975).

## 2.3 The Superbracket Decomposition

The inverse Jacobian matrix of a parallel manipulator has the Plücker coordinates of *6* lines in the projective space $\mathbb{P}^3$ as its rows. The superjoin of these *six* vectors in $\mathbb{P}^5$ corresponds to the determinant of their *six* Plücker coordinate vectors up to scalar multiple, which is the superbracket in GCA $\Lambda(V^{(2)})$ (White, 1983). Thus, a singularity occurs when these *six* Plücker coordinate vectors are dependent, which is equivalent to a superbracket equal to zero.

White, 1983 and McMillan, 1990, used the theory of projective invariants to decompose the superbracket into expression having brackets involving *12* points selected on the axes of these lines. Let [*ab,cd,ef,gh,ij,kl*] be the superbracket having six *2*-extensors as its column. These extensors represent lines *ab, cd, ef, gh, ij, kl* in the projective space, respectively. The expression of this superbracket taken from Ben-Horin and Shoham, 2006 is as follows

$$[ab,cd,ef,gh,ij,kl] = \left[abcd\right]\left[ef\overset{1}{g}\overset{2}{i}\right]\left[\overset{1}{h}\overset{2}{j}kl\right] - \left[ab\overset{3}{c}\overset{4}{e}\right]\left[\overset{3}{d}\overset{4}{f}gh\right]\left[ijkl\right] \\ - \left[ab\overset{5}{c}\overset{6}{e}\right]\left[\overset{5}{d}g\overset{7}{h}\overset{6}{i}\right]\left[\overset{7}{f}jkl\right] + \left[abc\overset{8}{g}\overset{9}{}\right]\left[\overset{8}{d}ef\overset{10}{i}\right]\left[\overset{9}{h}\overset{10}{j}kl\right] \quad (2)$$

where

$$\left[abcd\right]\left[ef\overset{1}{g}\overset{2}{i}\right]\left[\overset{1}{h}\overset{2}{j}kl\right]$$

denotes

$$\sum_{1,2}\mathrm{sign}(1,2)\left[abcd\right]\left[ef\overset{1}{g}\overset{2}{i}\right]\left[\overset{1}{h}\overset{2}{j}kl\right]$$

*1,2* are permutations of the *2*-element sets {*g, h*}, {*i, j*}, respectively.

Equation 2 may be transformed into a linear combination (sum) of *24* bracket monomials, where each bracket monomial is a product of 3 brackets. The monomials in Eq. 2 may be found in (McMillan, 1990 and Ben-Horin and Shoham, 2006).

## 3. Singularity Geometric Conditions

The aim of this study is to enlarge the application of line geometry to include *limited*-DOF manipulators whose legs apply actuation and constraint wrenches to the moving platform. These manipulators do not need to have a passive spherical joint anywhere along their legs and their 6x6-inverse Jacobian matrix may have Plücker coordinate vectors of finite and infinite lines as its rows. In this section, we use GCA and the superbracket decomposition to determine the singularity geometric conditions of two classes of parallel manipulators having inverse Jacobian matrices equivalent to those obtained for *3*-UPU translational and the parallel module of the Verne machine shown in Fig. 1 and 2.

## 3.1 Singularity analysis of the 3-U$\underline{P}$U translational manipulator

The *3*-UPU manipulator was studied in (Di Gregorio and Parenti-Castelli, 1998, Joshi and Tsai, 2002, Wolf and Shoham, 2003, Merlet, 2005). The moving platform controlled by three linear actuators along the three legs can perform a translational motion when the axes of the base universal joints are parallel to those of the platform universal joints of the same leg. Thus each leg *i* having connectivity equal to *5* applies one actuation force, $\hat{\mathbf{F}}_i = \begin{bmatrix} \mathbf{s}_i^T & (\mathbf{r}_i \times \mathbf{s}_i)^T \end{bmatrix}^T$, and one constraint moment, $\hat{\mathbf{M}}_i = \begin{bmatrix} \mathbf{0}_{1\times 3} & \mathbf{n}_i^T \end{bmatrix}^T$, to the moving platform, where $\mathbf{s}_i$ is a unit vector in the direction of the line of application of the force $\hat{\mathbf{F}}_i$ (leg *i*), $\mathbf{r}_i$ is the position vector of a point on this line and $\mathbf{n}_i$ is the cross product of the two vectors associated with the axes of the base U joint of leg *i* represented by that the direction of the torque associated with the constraint moment $\hat{\mathbf{M}}_i$. Each actuation force, $\hat{\mathbf{F}}_i$, is a zero pitch screw reciprocal to all joint screws (a joint screw stands for a twist screw associated with the joint) of leg *i* except for the joint screw associated with the actuated prismatic joint of the same leg. Each constraint moment, $\hat{\mathbf{M}}_i$, is an infinite pitch screw reciprocal to all joint screws of leg *i*. These actuation forces and constraint moments have Plücker coordinate vectors of finite and infinite lines, respectively, in the *3*-dimensional projective space. As a result, the 6x6- inverse Jacobian matrix will have the Plücker coordinate vectors of *3*-finite lines (actuation forces) and *3*-infinite lines (constraint moments) as its rows. The dependency between these lines is related to the degeneration of the inverse Jacobian matrix, which is equivalent to a superbracket equal to zero. The transpose of the inverse Jacobian of the 3-U$\underline{P}$U can be expressed as follows (Joshi and Tsai, 2002):

$$\left(J^{-1}\right)^T = \begin{bmatrix} \mathbf{s}_1 & \mathbf{s}_2 & \mathbf{s}_3 & \mathbf{0}_{1\times 3} & \mathbf{0}_{1\times 3} & \mathbf{0}_{1\times 3} \\ \mathbf{r}_1 \times \mathbf{s}_1 & \mathbf{r}_2 \times \mathbf{s}_2 & \mathbf{r}_3 \times \mathbf{s}_3 & \mathbf{n}_1 & \mathbf{n}_2 & \mathbf{n}_3 \end{bmatrix}$$
$$= \begin{bmatrix} \hat{\mathbf{F}}_1 & \hat{\mathbf{F}}_2 & \hat{\mathbf{F}}_3 & \hat{\mathbf{M}}_1 & \hat{\mathbf{M}}_2 & \hat{\mathbf{M}}_3 \end{bmatrix} \quad (3)$$

Let *ab*, *cd*, *ef*, be the finite lines representing the *3*-actuation forces $\mathbf{F}_i$ (*i=1, 2, 3*), where *a, b, c* are finite point and *b, d, f* are points at infinity.

All these points are expressed with their homogeneous coordinates, where $b = (\mathbf{s}_1^T \; 0)^T$, $d = (\mathbf{s}_2^T \; 0)^T$, $f = (\mathbf{s}_3^T \; 0)^T$. On the other hand, since every line at infinity meets every other line at infinity, the three constraint moments $\mathbf{M}_i$ ($i=1..3$) can be represented by three infinite lines $gh$, $gi$ and $hi$, respectively. According to Eq. 2 and due to the repetition of points in the same bracket, we simplify the superbracket expression into a reduced number of non-zero monomial terms and the superbracket decomposition of our manipulator, in turn, reduces to:

$$[ab, cd, ef, gh, gi, hi] = [abdf][cghi][eghi] \quad (4)$$

where $[abdf] = (\mathbf{s}_1 \times \mathbf{s}_2) \cdot \mathbf{s}_3$ and $[eghi] = [cghi] = (\mathbf{n}_1 \times \mathbf{n}_2) \cdot \mathbf{n}_3$

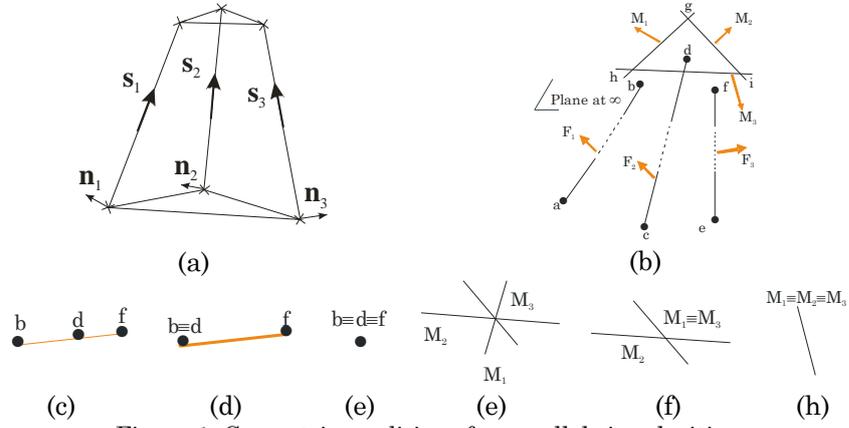

Figure 1: Geometric conditions for parallel singularities

Thus, the manipulator under study is at a singularity whenever vectors $\mathbf{s}_i$ or $\mathbf{n}_i$ ($i=1..3$) verify Eq. 4, which occurs respectively when triangle $ghi$ or $bdf$ vanishes (its surface is equal to zero). These two conditions include the cases (c-h) in Fig. 1. Each vertex of triangle $bdf$ is a point at infinity representing the direction of an actuation force. In case (c,d), the 3 points at infinity $b$, $d$ and $f$ are coincident, so three vectors $\mathbf{s}_i$ are coplanar. Case (d,e) occurs when at least two of the three points $b$, $d$, $f$ coincide and so at least two of the three forces are parallel. Each side of the triangle $ghi$ is a line at infinity defined as the intersection of a family of parallel planes at infinity, with $\mathbf{n}_i$ as the normal to these planes. Case (f, g) occurs when the three lines at infinity intersect at the same point, which means that three vectors $\mathbf{n}_i$ are coplanar. In case (g,h), at least two lines at infinity are collinear, which means that their corresponding planes are parallel and so at least two of the vectors $\mathbf{n}_i$ are parallel.

These conditions stand for all manipulators with an inverse Jacobian matrix consisting of three actuation forces and three constraint moments, like for example the *3*-U<u>P</u>U translational manipulator and the linear Delta robot (Clavel, 1988). Notice that another condition may appear when at least one constraint moment degenerates, meaning that at least one line *gh*, *gi* or *hi* degenerates to a point. This condition is more related to the arrangement of the joint within each leg. For the 3-U<u>P</u>U manipulator, this case can only happen if a universal joint is replaced by two revolute joints with parallel axes.

## 3.2   Singularity analysis of the Verne parallel module

The parallel module of the Verne machine consists of three legs, leg I, II and III (Fig. 2a). Each leg uses pairs of rods linking a prismatic joint to the moving platform through two pairs of spherical joints. Legs II and III are two identical parallelograms. Leg I differs from the other two legs in that *ac ≠ bd*. Leg I does not remain planar (rod directions define skew lines) as the machine moves, unlike what arises in the other two legs that are articulated parallelograms. The movement of the moving platform is generated by the slide of three actuators along three vertical guideways. We suppose that we are out of the serial singularities and we are interested only on studying the parallel singularities. We can thus consider that the transpose of the inverse Jacobian of this manipulator can be expressed as follow:

$$\left(\mathbf{J}^{-1}\right)^T = \begin{bmatrix} \hat{\mathbf{F}}_{11} & \hat{\mathbf{F}}_{12} & \hat{\mathbf{F}}_{21} & \hat{\mathbf{F}}_{22} & \hat{\mathbf{F}}_{31} & \hat{\mathbf{F}}_{32} \end{bmatrix} \tag{5}$$

The rows of $\mathbf{J}^{-1}$ are zero pitch screws, $\hat{\mathbf{F}}_{ij} = \begin{bmatrix} \mathbf{s}_{ij}^T & (\mathbf{r}_{ij} \times \mathbf{s}_{ij})^T \end{bmatrix}^T$ (*i=1..3, j=1, 2*), represented by finite lines along the six rods, where $\mathbf{s}_{ij}$ is a unit vector in the direction of the rod *j* within the leg *i* and $\mathbf{r}_{ij}$ is the position vector of a point on this rod. Each of those screws is an actuation force, which is reciprocal to all the joint screws of the rod *j* within the leg *i* except for the joint screw associated with the actuated prismatic joint of the same leg. Notice that the pair of actuation forces within legs II and III are parallel, so $\mathbf{s}_{21} = \mathbf{s}_{22}$ and $\mathbf{s}_{31} = \mathbf{s}_{32}$. The singularity geometrical conditions are associated with the dependency between these six actuation forces supported by the rods of the Verne parallel module.

Let *ab*, *cd*, *ef*, *gh*, *ij kl*, be respectively the finite lines representing the six actuation forces, where points *a*, *b*…,*l* are finite points located at the center of spherical joints (Fig. 2). We suppose that lines *ef*, *gh* are parallel and intersect at infinity as well as for lines *ij*, *kl*. Thus, the superbracket decomposition of these lines reduces to:

$$[ab, cd, ef, gh, ij, kl] = [am, cn, eo, go, ip, kp] \qquad (6)$$

where $m = b - a$, $n = d - c$, $ef \wedge gh = o = f - e = h - g$ and $ij \wedge kl = p = j - i = l - k$

The shortest form of the superbracket decomposition using the algorithm in (Ben-Horin et al., 2008) will result in the following none zero monomial terms:

$$[oegm][oncp][aipk] - [oega][oncp][mipk] - \\ [oegn][omap][cipk] + [oegc][omap][nipk] \qquad (7)$$

After collecting equal brackets and applying the shuffles relation (see White, 2005) caused by the permutation of *m*, *a* and *n*, *c*, we obtain:

$$[oncp][oeg\,\overline{m}][\overline{a}ipk] - [omap][oeg\,\overline{n}][\overline{c}ipk] = 0 \qquad (8)$$

Applying the following relation, $[(b-a)cde] = [bcde] - [acde]$, we are able to replace some points at infinity by finite points and obtain the following expression:

$$[ab, cd, ef, gh, ij, kl] = [oncp][feg\,\overline{b}][\overline{a}ijk] - [omap][feg\,\overline{d}][\overline{c}ijk] \qquad (9)$$

Using the meet operator relations (see White, 2005), we prove that:

$$[feg\,\overline{b}][\overline{a}ijk] = (ijk \wedge feg) \wedge ba = tu \wedge ba = [tuba] \text{ and} \\ [feg\,\overline{d}][\overline{c}ijk] = (ijk \wedge feg) \wedge dc = tu \wedge dc = [tudc] \qquad (10)$$

where *tu* is the line of intersection of leg II and III planes including such as: $\mathbf{tu} = (\mathbf{ef} \times \mathbf{eg}) \times (\mathbf{ij} \times \mathbf{ik})$

On other hand, it is easy to prove that:

$$[oncp] = [fepn] = [feqr] \text{ and } [omap] = [feqs] \qquad (11)$$

where $q = e + p$, $r = q + n$ and $s = q + m$

Finaly, the invariant algebraic expression related to the existence of parallel singularities of the parallel module of the Verne machine can be stated as:

$$[feqr][tuba] - [feqs][tudc] = 0 \qquad (12)$$

where $[tuba] = \mathbf{tu} \cdot (\mathbf{ub} \times \mathbf{ab})$, $[tudc] = \mathbf{tu} \cdot (\mathbf{ud} \times \mathbf{cd})$, $[feqr] = \mathbf{cd} \cdot \mathbf{N}$ and $[feqs] = \mathbf{ab} \cdot \mathbf{N}$ with $\mathbf{N} = \mathbf{ef} \times \mathbf{ij}$

The above expression is geometrically equivalent to the difference between the volume products of two pairs of tetrahedrons with vertices expressed as function of points *a,b,...l* as shown in Fig. 2c. This geometric condition includes the following cases: (*i*). planes of leg II and III are coplanar or parallel ($\|\mathbf{tu}\| = \mathbf{0}$); (*ii*). *ef* and *ij* are parallel ($\|\mathbf{N}\| = 0$); (*iii*). *ef*, *cd* are parallel and *ij*, *ab* are parallel or *ef*, *ab* are parallel and *ij*, *cd* are parallel; (*iv*). *ab* and *cd* intersect with *tu*, in this case the six actuation forces form a singular linear complex; (*v*) *ef*, *cd* are parallel and *tu*, *cd* are coplanar or *ij*, *cd* are parallel and *tu*, *cd* are coplanar or *ij*, *ab* are parallel and *tu*, *ab* are coplanar or *ef*, *ab* are parallel and *tu*, *ab* are coplanar; (*vi*). the *6* actuation forces form a general linear complex expressed by Eq 13.

$$[\mathbf{cd}\cdot\mathbf{N}]\left[\mathbf{tu}\cdot(\mathbf{ub}\times\mathbf{ab})\right] - [\mathbf{ab}\cdot\mathbf{N}]\left[\mathbf{tu}\cdot(\mathbf{ud}\times\mathbf{cd})\right] \qquad (13)$$

The singularity conditions of the Delta robot are obtained for the particular when $\mathbf{cd} = \mathbf{ab}$, thus Eq. 13 is reduced to:

$$[\mathbf{ab}\cdot\mathbf{N}]\left[\mathbf{tu}\cdot(\mathbf{db}\times\mathbf{ab})\right] = \left[(\mathbf{s}_1\times\mathbf{s}_2)\cdot\mathbf{s}_3\right]\left[(\mathbf{n}_I\times\mathbf{n}_2)\cdot\mathbf{n}_3\right] \qquad (14)$$

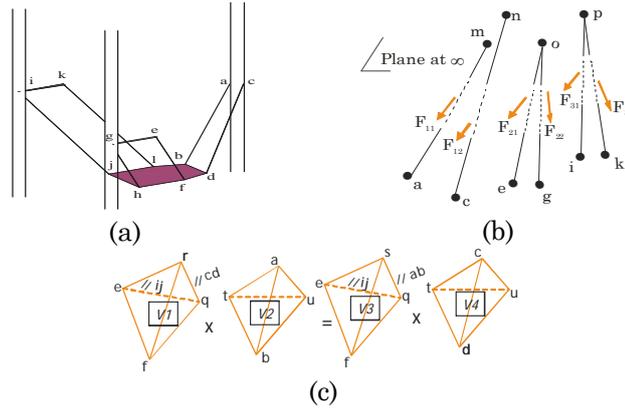

(a) (b)

(c)
Figure 2: Singularity Geometrical conditions of the Verne parallel module

## 6. Conclusions

This paper showed how GCA can be used to determine the geometric conditions associated with the singular configurations of limited-DOF parallel robots whose legs can transmit forces and moments to the moving platform. Three example robots were analyzed as illustrative examples. This method provides a physical meaning and a geometrical interpretation of singular configurations for a family of parallel manipulators, which is of interest for the designer of new robots.

## References

Ben-Horin, P. and Shoham, M. (2005), Singularity Analysis of Parallel Robots Based on Grassmann-Cayley Algebra, *International Workshop on Computational Kinematics*, Cassino, May 4-6.

Ben-Horin, P. and Shoham, M. (2006), Singularity condition of six degree-of freedom three-legged parallel robots based on Grassmann-Cayley algebra, *IEEE Transactions on Robotics*, no. 22, vol. 4, pp. 577-590.

Ben-Horin, P., Shoham, M., Caro, S., Chablat, D. and Wenger Ph. (2008), A Graphical user interface for the singularity analysis of parallel robots based on Grassmann-Cayley Algebra. *Advances in robot kinematics*.

Clavel, R. (1988), DELTA, A Fast Robot with Parallel Geometry, *Proceedings of 18th international symposium on industrial robots*, Lausanne, pp. 91-100.

Di Gregorio, R. and Parenti-Castelli, V. (1998), A translational 3-DOF parallel manipulator. *Advances in robot kinematics*, Lenarcic, J., Husty, M.L. (Eds.), Kluwer Academic Publishers, Dordrecht, pp. 49-58.

Joshi, SA and Tsai, LW (2002) Jacobian analysis of limited-DOF parallel manipulators. *ASME J Mech Des*, no. 124, pp. 254–258.

Hao, F. and McCarthy, J. (1998), Conditions for line-based singularities in spatial platform manipulators, *J. of Robotic Systems*, no. 15, vol. 1, pp. 43-55.

Kanaan, D., Wenger, Ph. and Chablat, D. (2006), Workspace Analysis of the Parallel Module of the VERNE Machine. *Problems of Mechanics*, no. 25, vol. 4, pp. 26-42.

McMillan, T. (1990), *Invariants of Antisymmetric Tensors*, PhD Dissertation, University of Florida.

Merlet, J.P. (1989), Singular configurations of parallel manipulators and Grassmann geometry, *International Journal of Robotics Research*, no. 8, vol. 5, pp. 45–56.

Merlet, J.P. (2006), *Parallel robots*, Second Edition, Springer.

Pottmann H., Peternell M., and Ravani B., (1999), An Introduction to Line Geometry with Applications, *Computer Aided Design*, vol. 31, pp. 3-16

Staffetti, E. and Thomas, F. (2000), Analysis of rigid body interactions for compliant motion tasks using the Grassmann-Cayley algebra, *Proceedings of the IEEE/RSJ International Conference on Intelligent Robotic System*.

White, N. (1975), The Bracket Ring of a Combinatorial Geometry I, *Transactions of the American Mathematical Society*, vol. 202, pp. 79-95.

White, N. (1983), The Bracket of 2-Extensors, *Congressus Numerantium*, vol. 40, pp. 419-428.

White, N. (1994), Grassmann-Cayley Algebra and Robotics, *Journal of Intelligent and Robotics Systems*, vol. 11, pp. 91-107.

White, N. (2005), Grassmann-Cayley Algebra and Robotics Applications, *Handbook of Geometric Computing*, part VIII pp. 629-656.

Wolf, A and Shoham, M (2003), Investigation of parallel manipulators using linear complex approximation, *ASME J Mech Des*, vol. 125, pp. 564–572

Zhao, T.-S., Dai, J. S. and Huang, Z. (2002), Geometric Analysis of Overconstrained Parallel Manipulators with Three and Four Degrees of Freedom, *JSME Int Journal. Ser C. Mech Systems, Mach Elem Manuf,* vol. 45, no. 3, pp.730-740